\documentclass{article}
\usepackage{spconf,amsmath,epsfig}
\usepackage{graphicx}
\usepackage{url}
\usepackage{tikz}
\usetikzlibrary{decorations.text}
\usepackage{array,multirow}
\usepackage{booktabs}
\usepackage{enumerate}

\title{Optimal transport vs. Fisher-Rao distance between copulas for clustering multivariate time series}
\name{G.~Marti$^{(1,2)}$, S.~Andler$^{(1,3)}$, F.~Nielsen$^{(2)}$, P.~Donnat$^{(1)}$}
 \address{$^{(1)}$Hellebore Capital Management, {\tt\small gautier.marti@helleborecapital.com}\\
	 $^{(2)}$ Ecole Polytechnique, {\tt\small nielsen@lix.polytechnique.fr}\\
	 $^{(3)}$ Ecole Normale Sup\'erieure de Lyon, {\tt\small sebastien.andler@ens-lyon.fr}}
\begin{document}
%
\maketitle
\begin{abstract}
We present a methodology for clustering N objects which are described by multivariate time series, i.e. several sequences of real-valued random variables. This clustering methodology leverages copulas which are distributions encoding the dependence structure between several random variables. 
To take fully into account the dependence information while clustering, we need a distance between copulas.
In this work, we compare renowned distances between distributions: the Fisher-Rao geodesic distance, related divergences and optimal transport, and discuss their advantages and disadvantages.
Applications of such methodology can be found in the clustering of financial assets. A tutorial, experiments and implementation for reproducible research can be found at \url{www.datagrapple.com/Tech}.
\end{abstract}

\section{Introduction}
\label{sec:introduction}

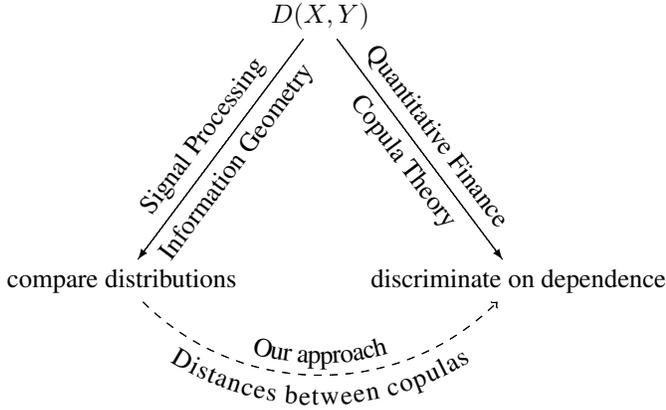
\begin{figure}
\begin{center}
\begin{tikzpicture}
[
    grow                    = down,
    sibling distance        = 15em,
    level distance          = 10em,
    edge from parent/.style = {draw, -latex},
    sloped
  ]
  \node {$D(X,Y)$}
    child { node (n1) {compare distributions}
    		edge from parent node [above] {Signal Processing}
    		edge from parent node [below] {Information Geometry} }
    child { node (n2) {discriminate on dependence}
    		edge from parent node [above] {Quantitative Finance}
    	    edge from parent node [below] {Copula Theory} } ;
    \def\myshift#1{\raisebox{1ex}}
    \draw [->,dashed,postaction={decorate,decoration={text along path,text align=center,text={|\myshift|Our approach}}}]      (n1) to [bend right=45] (n2);
    \def\myshift#1{\raisebox{-2.5ex}}
    \draw [->,dashed,postaction={decorate,decoration={text along path,text align=center,text={|\myshift|Distances between copulas}}}]      (n1) to [bend right=45] (n2);
    
\end{tikzpicture}\label{fig:approach}
\end{center}
\caption{Statistical distances from Information Geometry designed to compare distributions can help for clustering time series based on their dependence. Which ones are relevant for this task? And which ones are not?}
\end{figure}

What is clustering? Clustering is the task of grouping a set of objects in such a way that objects in the same group, also called cluster, are more similar to each other than those in different groups. Besides being hard to formalize and hard to solve, clustering essentially relies on the definition of a proper, but usually unknown, pairwise similarity measure between the data-points. Which class of distances should one use when the data-points are time series? Several surveys and experimental studies \cite{keogh2003need, liao2005clustering, aghabozorgi2015time} have tried to organize and benchmark this tentacular field of clustering time series: many application domains (e.g., astronomy, biology, energy, environment, medicine, robotics, speech recognition, finance) offer many goals that require specific distances and clustering approaches. When time series are considered as sequences of random variables, we can identify two main approaches: those who process signals by comparing their distributions \cite{dessein2013information,nielsen2013pattern} (Information Geometry), and those who measure the dependence (e.g., correlation, tail dependence) between them. In Econophysics, for example, Mantegna \textit{et al.} \cite{mantegna1999hierarchical} cluster stocks based on the pairwise correlation of their returns. 
Yet, we may want to represent an object by more than a single time series. For instance, a company could be represented by a $4$-variate time series consisting in (i) the returns of the stock, (ii) the traded stock volumes, (iii) the returns of its 5-year credit default swap (CDS) \cite{o2011modelling}, (iv) the traded CDS volumes.
Distances proposed so far \cite{yang2004pca,dasu2005grouping,singhal2005clustering} for clustering such multivariate time series do not aim specifically at discriminating on dependence. To this aim, Copula Theory provides convenient tools to encode the dependence between several time series as a distribution, the copula.
In order to cluster $N$ objects (for instance, companies) based on the dependence between their multivariate time series, we need a relevant distance between copulas.
Thus, our approach (depicted in Fig.~\ref{fig:approach}) is to leverage distances from Information Geometry to compare distributions - the copulas encoding dependence between the variates - in order to discriminate on the dependence between the random variables (but not on their distributions).
What kind of distances is relevant for comparing copulas? Far from being comprehensive, we illustrate our point with Wasserstein distances, Fisher-Rao geodesic distance and related divergences.

\section{Clustering methodology}

\subsection{A reminder on copulas}

Copulas are functions that couple multivariate distribution functions to their one-dimensional marginal distribution functions \cite{nelsen2013introduction}. This relationship is precised by Sklar's Theorem: 

\textbf{Theorem.} \textit{Sklar's Theorem \cite{sklar1959fonctions}.} For any random vectors $X = (X_1,\ldots,X_d)$ having continuous marginal cumulative distribution functions $P_i$, $1 \leq i \leq d$, its joint cumulative distribution $P$ is uniquely expressed as $P(X_1,\ldots,X_d) = C(P_1(X_1),\ldots,P_d(X_d))$, where $C$, the multivariate distribution of uniform marginals, is known as the copula of $X$.

Copulas are central for studying the dependence between random variables: their uniform marginals jointly encode all the dependence. They allow to study scale-free measures of dependence and are \textit{invariant to monotonous transformations of the variables}.

\subsection{Clustering multivariate time series using copulas}

We elaborate on a clustering methodology first described in \cite{marti2016optimal}.
Besides the associated paper, the basic methodology is illustrated at \url{www.datagrapple.com/Tech}.
Authors have suggested to:
\begin{itemize}
\item transform the data to its underlying copula,
\item compare this copula to other relevant copulas using the Earth Mover Distance \cite{rubner2000earth} (algorithmic version of a Wasserstein distance formulated as a linear problem and solved with the Hungarian algorithm),
\item use any appropriate clustering algorithm that can handle a dissimilarity matrix as input.
\end{itemize}
The methodology presented in \cite{marti2016optimal} has several \textit{advantages}: non-parametric, robust to noise and deterministic (properties of the Earth Mover Distance); non-parametric, robust to noise, fast converging, accurate and generic representation of dependence (properties of empirical copulas).
But the approach has also some serious \textit{scaling issues}: (i) in dimension, non-parametric estimations of density suffer from the curse of dimensionality, (ii) in time, the Earth Mover Distance is costly to compute. 

In order to alleviate the previous drawbacks, we may consider a parametric version of this clustering methodology. It implies the choice of a parametric copula (e.g., Gaussian copula, Student-t copula, Archimedean copula) and the choice of a statistical distance to compare the copulas.
What is a relevant distance to measure the resemblance of copulas?

\section{Sensitivity of distances with respect to dependence}

\subsection{A reminder on statistical distances}

Statistical distances are distances between probability distributions.
Many such distances have been designed to deal with practical problems in signal processing \cite{nielsen2015geometric}.

One of the leading approaches is to consider the parameter space $\Theta = \{ \theta \in \mathbf{R}^D ~|~ \int p_{\theta}(x)dx = 1 \}$ of a
family of parametric probability distributions $\{p_{\theta}(x)\}_{\theta}$ with $x \in \mathbf{R}^d$ and $\theta \in \mathbf{R}^D$ as a  Riemannian manifold endowed with the Fisher-Rao metric $ds^2(\theta) = \sum_{i=1}^D \sum_{j=1}^D g_{ij}(\theta)d\theta_i d\theta_j$ \cite{rao1992information}. The coefficients $g_{ij}(\theta) = \mathbf{E}_{\theta}\left[ \frac{1}{p(\theta)} \frac{\partial p}{\partial \theta_i} \frac{1}{p(\theta)} \frac{\partial p}{\partial \theta_j} \right] = g_{ji}(\theta)$ are known as the Fisher Information Matrix coefficients.
Two probability distributions represented by their respective density $p_{\theta_1}$ and $p_{\theta_2}$ are considered as two points $\theta_1$ and $\theta_2$ on the manifold $(\Theta, ds^2)$.
The Fisher-Rao geodesic distance between these two probability distributions can be computed by integrating the Fisher-Rao metric along the geodesics (locally shortest paths) between the corresponding points $\theta_1$ and $\theta_2$: $D(\theta_1,\theta_2) = \int_{\theta_1}^{\theta_2} ds$.

Since computing geodesics which requires solving ordinary differential equations (ODEs) can be intractable, one often considers related divergences such as Kullback-Leibler, symmetrized Jeffreys, Hellinger, or Bhattacharrya divergences which coincide with the quadratic form approximations of the Fisher-Rao distance between two close distributions, and which are computationally more tractable. These divergences all belong to the class of Ali-Silvey-Csisz\'ar $f$-divergences, enjoy the information monotonicity \cite{amari2009divergence} (coarsing bins decrease the divergence value), are \textit{invariant under reparametrizations}, and furthermore induce the $\pm\alpha$-geometry for $\alpha=3+2f'''(1)$ (where $f$ is a convex function).

Alternatively to the Fisher-Rao geometry, Wasserstein distances \cite{villani2008optimal} provide another natural way to compare probability distributions. 
Given a metric space $M$, these distances optimally transport the probability measure $\mu$ defined on $M$ to turn it into $\nu$:
$$W_{p} (\mu, \nu):=\left( \inf_{\gamma \in \Gamma (\mu, \nu)} \int_{M \times M} d(x, y)^{p} \, \mathrm{d} \gamma (x, y) \right)^{1/p},$$
where $p \geq 1$, $\Gamma(\mu,\nu)$ denotes the collection of all measures on $M \times M$ with marginals $\mu$ and $\nu$. 
It can be observed that \textit{computing the Wasserstein distance between two probability measures amounts to finding the most correlated copula associated with these measures.} Notice also that unlike Fisher-Rao and related divergences, Wasserstein distances work with probability measures instead of probability density functions.

\subsection{Distances between Gaussian copulas}

We illustrate the behaviour of these distances in the simple case where the underlying copula is a Gaussian (which may not be relevant for all applications). Moreover, when the compared distributions are multivariate Gaussians, we have analytical formulas (which are reported in Table~\ref{tab:distances}).

\begin{figure}
\begin{center}
\includegraphics[width=0.32\linewidth]{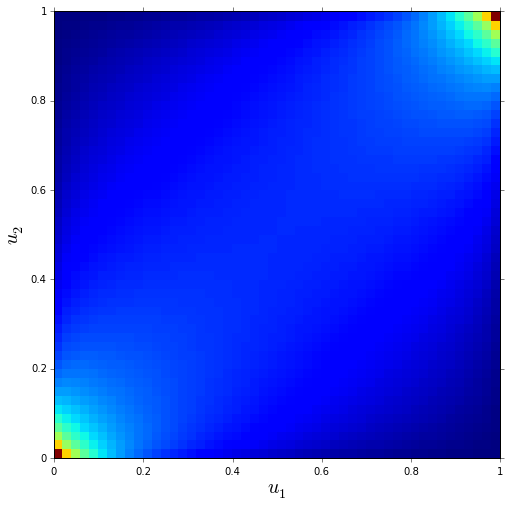}
\includegraphics[width=0.32\linewidth]{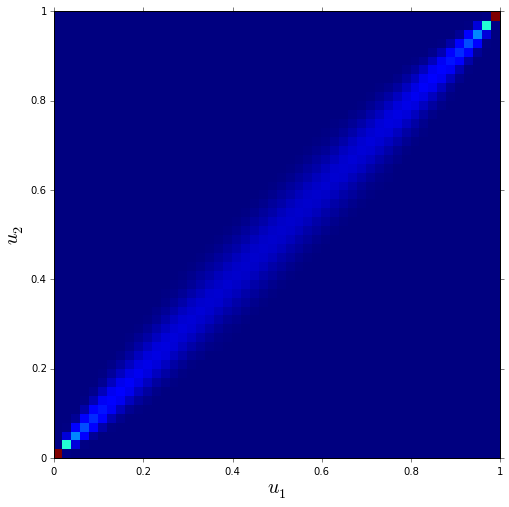}
\includegraphics[width=0.32\linewidth]{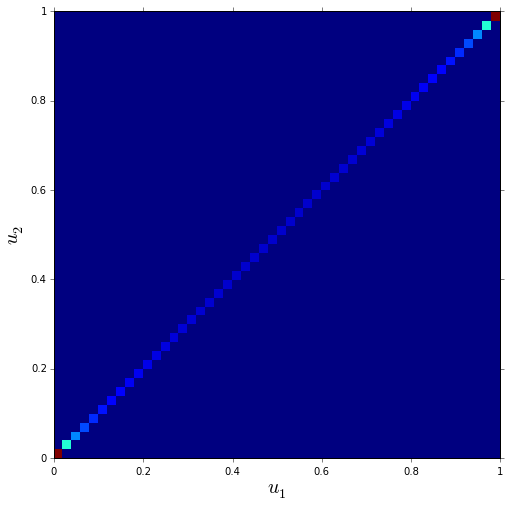}
\end{center}
\caption{Densities of $C_{R_A}^{\mathrm{Gauss}}, C_{R_B}^{\mathrm{Gauss}}, C_{R_C}^{\mathrm{Gauss}}$ respectively;  Notice that for strong correlations, the density tends to be distributed very close to the diagonal.}
\label{fig:copula_pdf}
\end{figure}

The Gaussian copula is a distribution over the unit cube $[0,1]^d$.
It is constructed from a multivariate normal distribution over $\mathbf{R}^d$
by using the probability integral transform.
For a given correlation matrix $R \in \mathbf{R}^{d\times d}$, the Gaussian copula with parameter matrix $R$ can be written as
$$C_R^{\mathrm{Gauss}}(u_1,\ldots,u_d) = \Phi_R(\Phi^{-1}(u_1),\ldots,\Phi^{-1}(u_d)),$$
where $\Phi^{-1}$ is the inverse cumulative distribution function of a standard normal and $\Phi_R$ is the joint cumulative distribution function of a multivariate normal distribution with mean vector zero and covariance matrix equal to the correlation matrix $R$. For illustration purposes, we consider three bivariate Gaussian copulas parameterized by $$R_A = \begin{pmatrix}
   1 & 0.5 \\
   0.5 & 1 
\end{pmatrix}, R_B = \begin{pmatrix}
   1 & 0.99 \\
   0.99 & 1 
\end{pmatrix},$$ and $R_C = \begin{pmatrix}
   1 & 0.9999 \\
   0.9999 & 1 
\end{pmatrix}$ respectively. Heatmaps of their densities are plotted in Fig.~\ref{fig:copula_pdf}.

\begin{table*}
\caption{Distances in closed-form between Gaussians and their sensitivity to the correlation strength}
\begin{center}
\begin{tabular}{ccccc}
    \toprule
           & $D\left(\mathcal{N}(0,\Sigma_1),\mathcal{N}(0,\Sigma_2)\right)$ & $D(R_A,R_B)$ & & $D(R_B,R_C)$ \\ \midrule
    Fisher-Rao \cite{atkinson1981rao} & $\sqrt{ \frac{1}{2}\sum_{i=1}^n (\log \lambda_i)^2 }$  & 2.77 & $<$ & 3.26 \\
    $KL(\Sigma_1||\Sigma_2)$ & $\frac{1}{2} \left( \log \frac{|\Sigma_2|}{|\Sigma_1|} - n + tr(\Sigma_2^{-1} \Sigma_1) \right) $ & 22.6 & $<$ & 47.2 \\
    Jeffreys & $KL(\Sigma_1||\Sigma_2) + KL(\Sigma_2||\Sigma_1)$  & 24 & $<$ & 49    \\
    Hellinger & $\sqrt{ 1 - \frac{|\Sigma_1|^{1/4} |\Sigma_2|^{1/4}}{|\Sigma|^{1/2}} }$ & 0.48 & $<$ & 0.56    \\
    Bhattacharyya & $\frac{1}{2} \log \frac{|\Sigma|}{\sqrt{|\Sigma_1| |\Sigma_2| } } $   & 0.65 & $<$ & 0.81    \\
    $W_2$ \cite{takatsu2011wasserstein} & $\sqrt{ tr\left( \Sigma_1 + \Sigma_2 - 2\sqrt{\Sigma_1^{1/2} \Sigma_2 \Sigma_1^{1/2}} \right) }$ & \textbf{0.63} & $\mathbf{>}$ & \textbf{0.09}   \\ \bottomrule
  \end{tabular}
  
  $\lambda_i$ eigenvalues of $\Sigma_1^{-1}\Sigma_2$; $\Sigma = \frac{\Sigma_1 + \Sigma_2}{2}$
\end{center}\label{tab:distances}
\end{table*}

\subsection{Distance sensitivity: An unwanted property for clustering copulas?}

In Table~\ref{tab:distances}, we report the distances $D(R_A,R_B)$ between $C_{R_A}^{\mathrm{Gauss}}$ and $C_{R_B}^{\mathrm{Gauss}}$, and the distances $D(R_B,R_C)$ between $C_{R_B}^{\mathrm{Gauss}}$ and $C_{R_C}^{\mathrm{Gauss}}$.
We can observe that unlike Wasserstein $W_2$ distance, Fisher-Rao and related divergences consider that $C_{R_A}^{\mathrm{Gauss}}$ and $C_{R_B}^{\mathrm{Gauss}}$ are nearer than $C_{R_B}^{\mathrm{Gauss}}$ and $C_{R_C}^{\mathrm{Gauss}}$.
This may sound surprising since $C_{R_B}^{\mathrm{Gauss}}$ and $C_{R_C}^{\mathrm{Gauss}}$ both describe a strong positive dependence between the two variates whereas $C_{R_A}^{\mathrm{Gauss}}$ describes only a mild positive dependence.

Our geometric intuition to explain this fact is that Fisher-Rao geodesic distance and its related divergences are only defined on the manifold of probability distribution densities. However, the copula characterizing comonotonicity (perfect positive dependence), known as the Fr\'echet-Hoeffding upper bound copula $M(u_1,\ldots,u_d) = \min \{u_1,\ldots,u_d\}$, has no density. So, perfect positive dependence (for a bivariate Gaussian, it means that the two variates are perfectly correlated: $\rho = 1$) is not a point of the manifold.
Unlike these distances, Wasserstein distances are defined between probability measures, so no such problem arises for the Fr\'echet-Hoeffding upper bound copula.
In the Gaussian case considered, the closed-form formulas for these distances can make this intuition clearer. For Fisher-Rao and related divergences, distances are expressed using the inverse of the covariance matrix and the inverse of its determinant. These matrices are ill-conditioned when correlation is strong, and singular when correlation is perfect. For Wasserstein $W_2$ distance, the formula is well defined in terms of square roots.

\begin{figure}
\begin{center}
\includegraphics[width=0.23\linewidth]{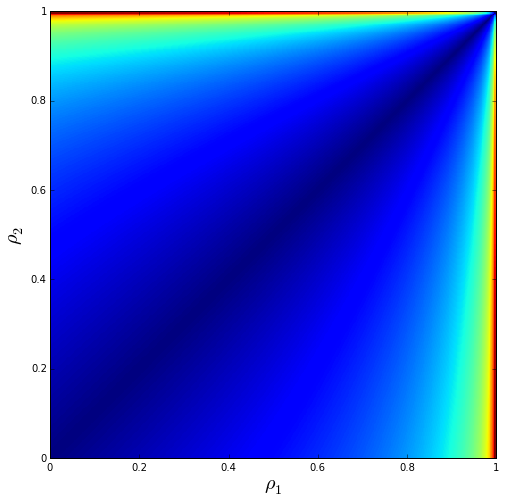}
\includegraphics[width=0.24\linewidth]{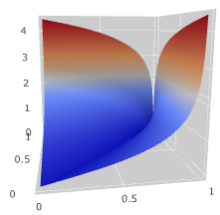}
\includegraphics[width=0.23\linewidth]{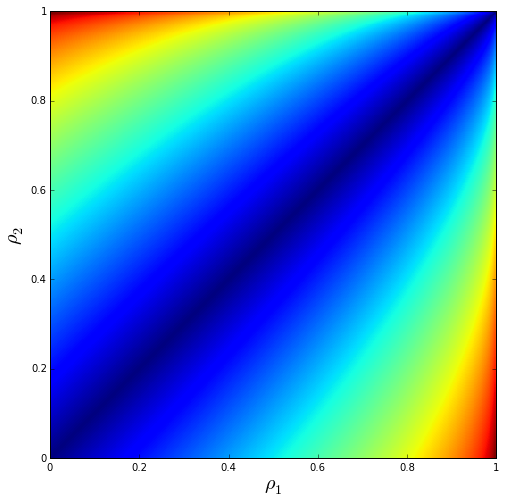}
\includegraphics[width=0.23\linewidth]{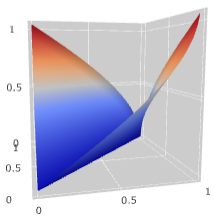}
\end{center}
\caption{Distance heatmap and surface as a function of $(\rho_1,\rho_2)$ for Fisher-Rao (left), for Wasserstein $W_2$ (right)}\label{fig:rao_dist}
\end{figure}

In \cite{barbaresco2011geometric}, Barbaresco gives an extensive comparison of Fisher-Rao geometry versus Wasserstein geometry on the space of covariance matrices. One of the noticeable difference is that the Fisher-Rao geometry has negative curvature whereas Wasserstein geometry is flat and has nonnegative curvature. 
The notion of curvature is key to understand the behaviour of clustering using statistical distances.
For instance, we have displayed in Fig.~\ref{fig:rao_dist} the distances $D(\rho_1,\rho_2)$ between the Gaussian copulas parameterized by $$\begin{pmatrix}
   1 & \rho_1 \\
   \rho_1 & 1 
\end{pmatrix} \textrm{~and~} \begin{pmatrix}
   1 & \rho_2 \\
   \rho_2 & 1 
\end{pmatrix}.$$
One can notice that Wasserstein $W_2$ exhibits a roughly linear increase away from the diagonal with low curvature: It can discriminate equally well for all parameters $(\rho_1,\rho_2) \in [0,1]^2$. On the contrary, the behaviour of Fisher-Rao strongly depends on the values $\rho_1, \rho_2$ as shown in Fig.~\ref{fig:rao_dist}: For high correlations, a small change induces a big change on the distance value due to the curvature. We call it sensitivity.
In addition to returning counter-intuitive distance values as reported in Table~\ref{tab:distances}, this property could lead to totally spurious distances and thus clusters when working with finite sample data: What if the parameters estimation error is bigger than the sensitivity? In practice, the distance would be useless. 

However, Fisher-Rao and related divergences do not suffer from this drawback. They all can be locally written as a quadratic form of the Fisher Information Matrix $\mathcal{I}(\rho)$. Through this connection to the Cram\'er-Rao Lower Bound $\mathrm{Var}(\hat{\rho}) \geq \frac{1}{\mathcal{I}(\rho)}$ \cite{rao1992information}, they deviate (the distance sensitivity) just the right amount with respect to the statistical uncertainty of the estimator. For Pearson correlation estimate, we have $\mathrm{Var}(\hat{\rho}) \geq \frac{1}{\mathcal{I}(\rho)} = \frac{(\rho-1)^2(\rho+1)^2}{3(\rho^2+1)}$, i.e. stronger the correlation, finer the estimate can be.



\section{Clustering experiments}

\begin{figure}
\begin{center}
\includegraphics[width=0.15\linewidth]{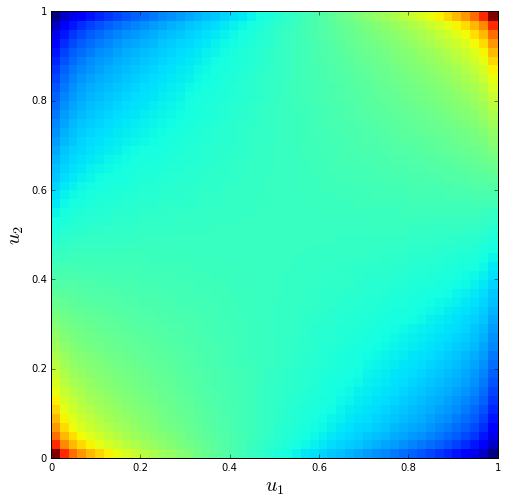}
\includegraphics[width=0.15\linewidth]{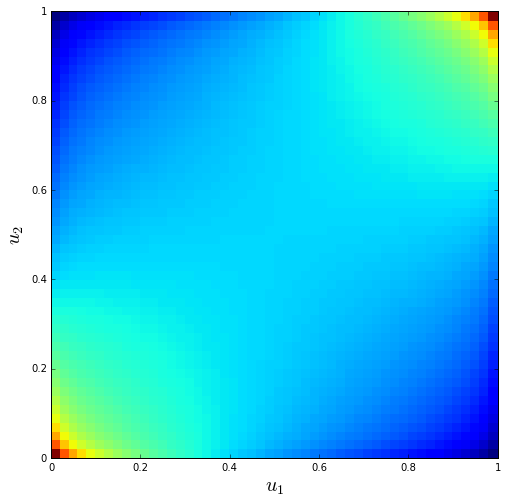}
\includegraphics[width=0.15\linewidth]{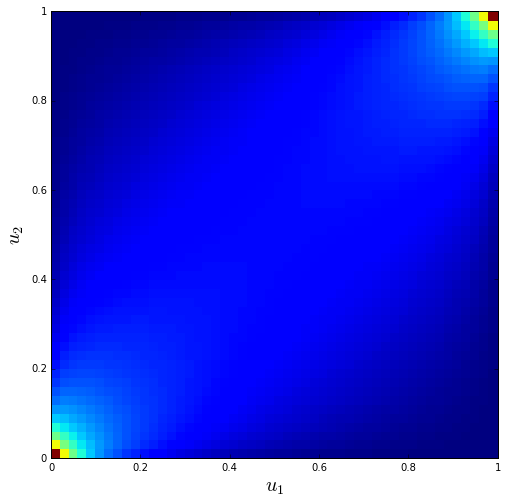}
\includegraphics[width=0.15\linewidth]{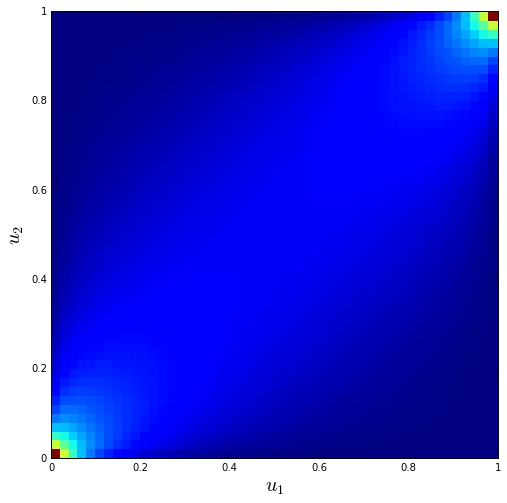}
\includegraphics[width=0.15\linewidth]{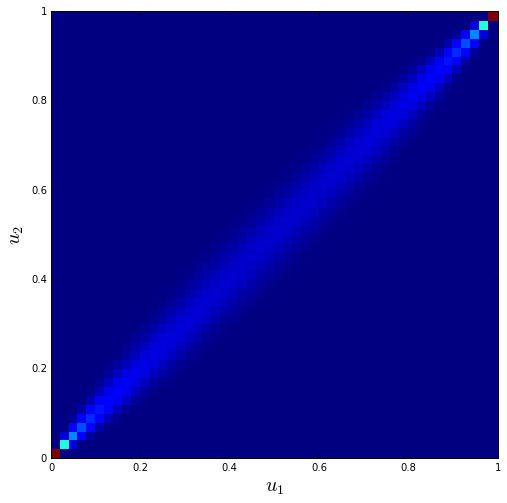}
\includegraphics[width=0.15\linewidth]{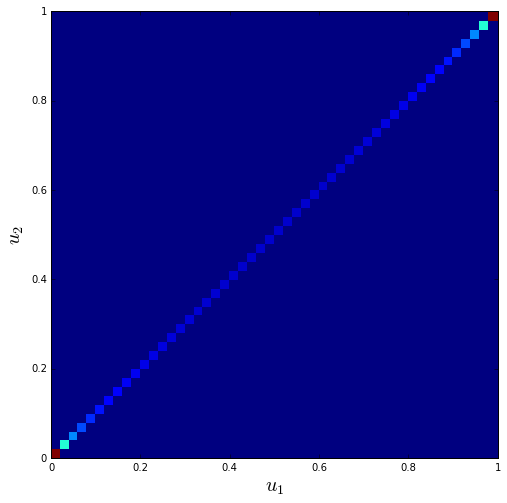}
\end{center}
\caption{Datasets of bivariate time series are generated from six Gaussian copulas with correlation .1, .2, .6, .7, .99, .9999}\label{fig:correl_clusters}
\end{figure}

\begin{figure}
\begin{center}
\includegraphics[width=0.46\linewidth]{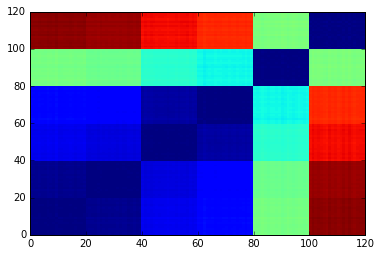}
\includegraphics[width=0.46\linewidth]{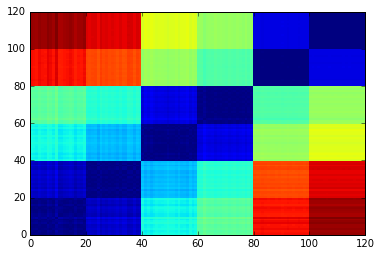}
\end{center}
\caption{Distance heatmaps for Fisher-Rao (left), $W_2$ (right); Using Ward clustering, Fisher-Rao yields clusters of copulas with correlations $\{.1,.2,.6,.7\}$, $\{.99\}$, $\{.9999\}$, $W_2$ yields $\{.1,.2 \}$, $\{.6,.7 \}$, $\{.99,.9999 \}$ }\label{fig:dist_clusters}
\end{figure}

Fisher-Rao geodesic distance has been successfully applied for clustering and classification \cite{el2011color}, statistical analysis (e.g., mean, median, PCA) on covariance manifolds in computational anatomy \cite{pennec2009statistical} and radar processing \cite{barbaresco2011geometric}. In financial applications, variates tend to be strongly correlated (for instance, correlation between maturities in a term structure can be up to 0.99). In such cases, the sensitivity problem discussed above may impair the clustering results.
We illustrate this assertion by considering a dataset of $N$ bivariate time series evenly generated from the six Gaussian copulas depicted in Fig.~\ref{fig:correl_clusters}. When a clustering algorithm such as Ward is given a distance matrix computed from Fisher-Rao (displayed in Fig.~\ref{fig:dist_clusters}), it will tend to gather in a cluster all copulas but the ones describing high dependence which are isolated. $W_2$ yields a more balanced and intuitive clustering where clusters contain copulas of similar dependence.
Code for the numerical and clustering experiments are available at \url{www.datagrapple.com/Tech}.


\section{Discussion}

In this paper, we have focused on Gaussian copulas for two reasons: (i) we know closed-form formulas for the distances between multivariate Gaussian distributions; (ii) the existing machine learning literature focus on the manifold of covariances \cite{abou2012note}. We have shown that if the dependence is strong between the time series, the use of Fisher-Rao geodesic distance and related divergences may not be appropriate. 
They are relevant to find which samples were generated from the same set of parameters (clustering viewed as a generalization of the three-sample problem \cite{Ryabko:10clust}) due to their local expression as a quadratic form of the Fisher Information Matrix determining the Cram\'er-Rao Lower Bound on the variance of estimators. To measure distance between copulas, we think that the Wasserstein geometry is more appropriate since it does not lead to these counter-intuitive clusters.
Beyond the Gaussian case, the phenomenon illustrated here should subsist as Fisher-Rao is defined on manifold of densities but the copula for comonotonicity cannot be part of it.
We will investigate further this issue. We would also like to encompass the embedding of probability distributions into reproducing kernel Hilbert spaces \cite{sriperumbudur2009kernel} in our comparison of the possible distances for copulas.


\bibliographystyle{IEEEbib}
\bibliography{refs}

\end{document}